\documentclass[runningheads]{llncs}

\usepackage[mobile]{eccv}

\usepackage{eccvabbrv}
\usepackage[detect-all,per-mode=symbol,group-digits=integer,group-minimum-digits=4,group-separator={,}]{siunitx}

\usepackage{graphicx}
\usepackage{booktabs}

\usepackage[accsupp]{axessibility}  %

\usepackage{hyperref}

\usepackage{orcidlink}

\usepackage{pgfplots}
\pgfplotsset{width=7cm,compat=1.8}
\usepackage{pgfplotstable}

\usepackage{amsmath}

\newenvironment{smallpmatrix}
  {\left(\begin{smallmatrix}}
  {\end{smallmatrix}\right)}

\begin{document}

\title{Affine steerers for structured keypoint description}

\author{Georg Bökman$^1$ \and Johan Edstedt$^2$ \and Michael Felsberg$^2$ \and Fredrik Kahl$^1$}

\authorrunning{G.~Bökman et al.}

\institute{Chalmers University of Technology \and Linköping University}

\maketitle

\begin{abstract}
We propose a way to train deep learning based keypoint descriptors that makes them approximately equivariant for locally affine transformations of the image plane.
The main idea is to use the representation theory of GL(2) to generalize the recently introduced concept of steerers from rotations to affine transformations. Affine steerers give high control over how keypoint descriptions transform under image transformations.
We demonstrate the potential of using this control for image matching.
Finally, we propose a way to finetune keypoint descriptors with a set of steerers on upright images and obtain 
state-of-the-art results on several standard benchmarks.
Code will be published at \href{https://github.com/georg-bn/affine-steerers}{github.com/georg-bn/affine-steerers}.
  \keywords{Keypoint description \and Image matching \and Equivariance}
\end{abstract}

\section{Introduction}
\label{sec:intro}
Image matching is a critical component in a wide range of computer vision applications, including 3D scene reconstruction, stereo imaging, and motion tracking. Initially, the field advanced through the development of sophisticated engineering methods, tailored to achieve robust matching capable of handling significant changes in image scale and rotation, while being resilient to variations in lighting and different camera viewpoints. However, these manually crafted approaches have gradually been superseded by deep learning techniques, which offer a more adaptable and powerful framework for tackling the many other complexities of image matching. Unfortunately, these methods lost much of the robustness of the handcrafted approach. In this paper, we aim to further enhance the performance of deep learning based image matching by introducing a method specifically designed to improve robustness to affine distortions. This approach not only capitalizes on the strengths of deep learning but also revisits and revitalizes a foundational concept from early research in the field: ensuring robustness to affine distortions is essential for reliable image matching across wide baselines.

\begin{figure}
    \centering
    \includegraphics[width=\textwidth]{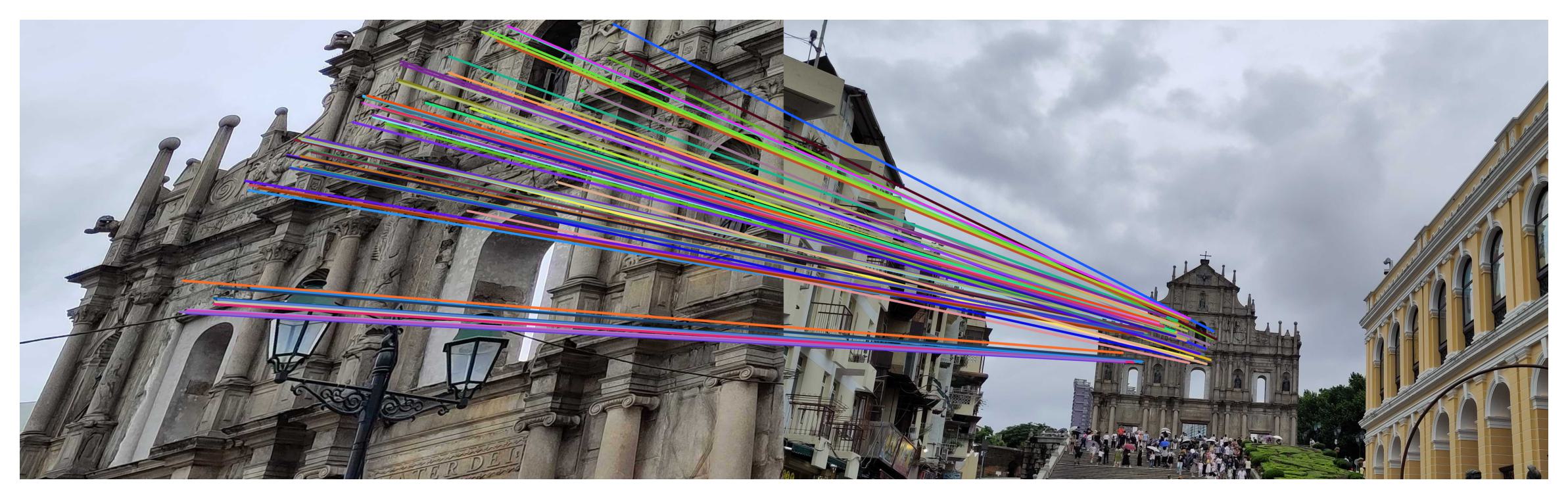}
    \caption{\textbf{Qualitative example.} We show qualitative matching results of our \emph{AffSteer-B} descriptor which achieves SotA results for detector-descriptors on several benchmarks for image matching. We plot inliers after homography estimation with RANSAC.}
    \label{fig:teaser}
\end{figure}

Our approach hinges on training neural network based
keypoint descriptors that are approximately 
equivariant under local affine transformations. We train these networks
by generalizing the \emph{steerers} framework \cite{bökman2024steerers}
from $\mathrm{SO}(2)$ to $\mathrm{GL}(2)$.
We also find that training with affine steerers with heavy
homography augmentation
works well as pretraining before fine-tuning on upright images.
This approach, while sacrificing some degree of equivariance,
achieves new SotA results on standard benchmarks for upright image matching.
In particular, we improve on IMC22~\cite{image-matching-challenge-2022}
from 72.9~mAA@10 achieved by DeDoDe-B~\cite{edstedt2024dedode} to
77.3~mAA@10 with our model \emph{AffSteer-B}.
A qualitative example on images taken by the authors is shown in Figure~\ref{fig:teaser}.
We also obtain
competitive results %
on the rotation variant benchmark AIMS~\cite{stoken2023astronaut} by using our equivariant model \emph{AffEqui-B}.
Finally, we analyze the equivariance properties of our descriptors
and outline promising directions for future research.
In summary, our main contributions are:
\begin{enumerate}
    \item Generalizing the steerers concept to image transformations that can be locally approximated by the affine group (Sections~\ref{sec:background} and \ref{sec:gl2}).
    \item Describing how to train steerers for the affine group (Section~\ref{sec:train-megadepth}).
    \item Introducing a new training procedure for upright-specialized descriptors
    consisting of first pretraining with affine steerers and then fine-tuning on upright images through the max similarity method (Sections~\ref{sec:pretrain}, \ref{sec:finetune}).
    \item Evaluating our methods on a wide range of standard benchmarks, with
     SotA results for detector-descriptor based methods (Section~\ref{sec:experiments}).
    \item Critically examining the properties of our descriptors
    (Sections~\ref{sec:steer_cap}, \ref{sec:conclusion}).
\end{enumerate}

\begin{figure}
    \centering
    \includegraphics[width=\textwidth]{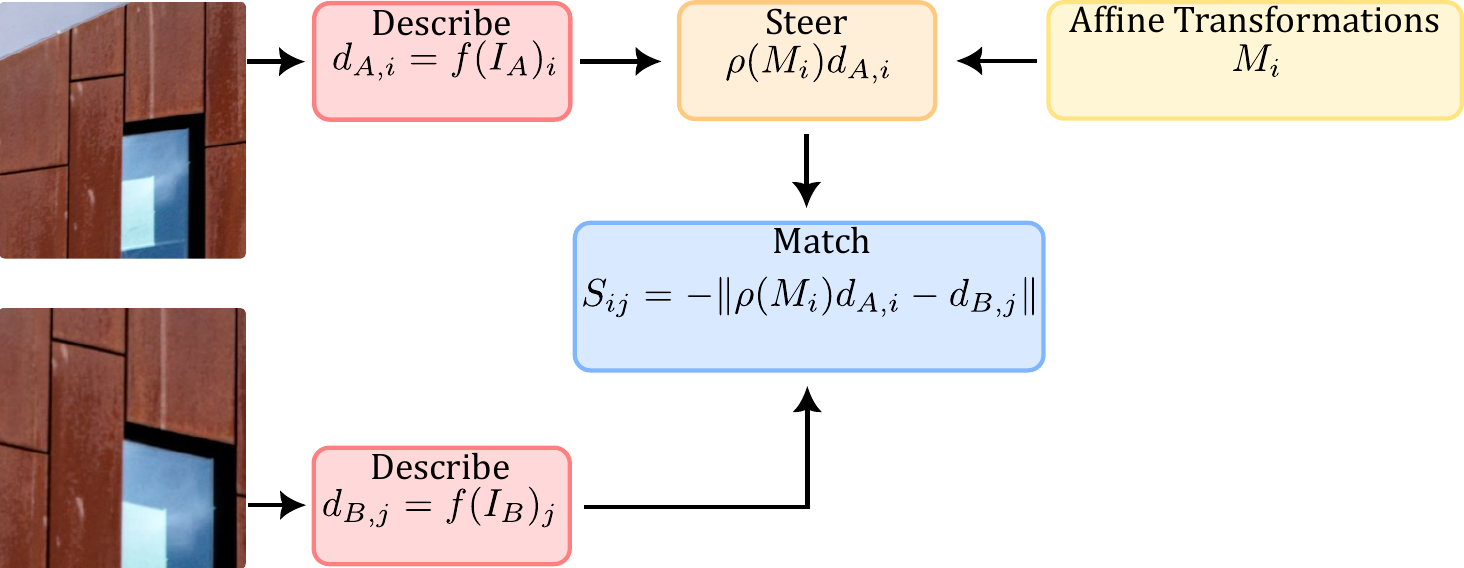}
    \caption{\textbf{Overview of the steering idea for keypoint matching.}
    An affine steerer gives a way to modify descriptions as if they were obtained
    from warped images, without having to rerun the descriptor network on warped images. The steerer is a linear map and hence computationally light to use.
    }
    \label{fig:overview}
\end{figure}

\subsection{Related work}
\paragraph{Image matching.}
Image matching has classically been approached by detector and descriptor based methods~\cite{lowe2004distinctive,yi2016lift,revaud2019r2d2,dusmanu2019d2,detone2018superpoint, tyszkiewicz2020disk,gleize2023silk,zhao2022alike,Zhao2023ALIKED} that detect keypoints~\cite{barroso2019key, tian2020hynet,yan2022learning,TUZNIK_OLVER_TANNENBAUM_2020} and compute descriptions~\cite{lee2021self,tian2019sosnet,mishkin2018repeatability,luo2019contextdesc,mishchuk2017working, balntas2016learning, jonsson2009efficient} typically matched using mutual nearest neighbors (MNNs). It was early on recognized that robustness to affine distortions plays an important role in order to achieve good performance \cite{matas2004robust,mikolajczyk-ijcv-2004,mikolajczyk-pami-2005,forssen2007shape,olverFeatureMatchingHeat2020}. ASIFT, which extends the Scale Invariant Feature Transform (SIFT) algorithm \cite{lowe2004distinctive} by test time augmentation to offer affine invariance, was introduced in \cite{asift-2011}. 
More recently, sparse keypoint matchers such as SuperGlue~\cite{sarlin20superglue} and later works~\cite{lindenberger2023lightglue,chen2021learning,shi2022clustergnn} aim to improve on the simple descriptor MNN matching using graph neural networks. Detector-free methods instead match on a coarse regular grid followed by sparse refinement with notable examples such as LoFTR~\cite{sun2021loftr}, and numerous following works~\cite{wang2022matchformer,huang2023adaptive,yuandchang2023astr,cao2023casmtr,giang2022topicfm,bokman2022case,mao20223dg}. Finally, 
dense methods that aim to estimate matches on a fine regular grid such as GLU-Net~\cite{truong2020glu} have lately received increased interest~\cite{truong2020gocor,truong2021learning,truong2023pdc,edstedt2023dkm,edstedt2024roma,ni2023pats,zhu2023pmatch}.
In this paper we focus on sparse keypoint descriptors that rely on MNN matching. This is the workhorse for numerous applications in computer vision. We use the SotA keypoint descriptor DeDoDe~\cite{edstedt2024dedode} as our baseline.

\paragraph{Affine correspondences.} Our work is also related to methods that rely on affine correspondences. That is, apart from the keypoint location, also the affine transformation that maps one local neighborhood region to the corresponding one in the other image, is used in downstream tasks such as estimating the epipolar geometry \cite{matas2002local,bentolila-cviu-2014,barath-eccv-2020,barath2023affineglue}. There are also learned methods that provide an affine canonical frame around each keypoint, for instance, AffNet \cite{affnet-eccv-2018} which is important, \eg, in image retrieval~\cite{obdrvzalek2002local}. %

\paragraph{Equivariant representation learning.}
Self-supervised visual representations~\cite{hadsell2006dimensionality, he2022mae, oquab2023dinov2,zhou2022ibot} are typically learned by maximizing agreement between two views~\cite{chen2021learning, caron2021dino, grill2020bootstrap}, which is reminiscent of the image matching task, where we seek to learn local visual representations that match across views.
Recent works~\cite{cohen2014transformation, koyama2024neural, shakerinava2022structuring, gupta2023structuring, garrido2023sie, garrido2024worldmodel, pmlr-v162-melnyk22a, park2022learning} investigate explicitly learning equivariant representations, which makes the learned representations structured in the sense that it is possible to interpret certain changes in the latent space as specific changes to the input images.
The equivariance can also be hard-coded into the network architecture,
\eg, for compact groups such as the rotation group \cite{cohen2016group, weiler2019general, zznet}. This approach has been shown to work well for rotation invariant keypoint detection~\cite{han2021redet,lee2022self,santellani2023strek} and description~\cite{bokman2022case,lee2023learning}.
For the affine group acting on images, hard-coding equivariance has only been
done approximately via sampling on the group and with
small networks~\cite{MacDonald_2022_CVPR, mironenco2024lie},
prohibiting the use of such approaches for SotA keypoint description.
Unrelated to images, networks equivariant under the group $\mathrm{SL}(2)$ were considered for polynomial optimization
problems in \cite{lawrence2023learning}, using
the same irreps from group representations on homogeneous polynomials that we will use
for the group $\mathrm{GL}(2)$.
In sharp contrast to hard-coding equivariance, it has been found that even without explicitly encouraging equivariance in learned representations, it often appears approximately because of symmetries in the data~\cite{lenc2015understanding, gruver2023the, bökman2023investigating, Bruintjes_2023_CVPR, bökman2024steerers}.
Closest to our work, Bökman \etal~\cite{bökman2024steerers} propose a framework for learning approximately rotationally equivariant keypoint descriptors. We build on that work, generalizing it to the larger group of $\mathrm{GL}(2)$.

\nocite{brynte2024}

\section{Background}\label{sec:background}

The primary goal of keypoint matching is to identify 2D points across two images of the same scene, representing the same 3D points. A pair of
corresponding 2D points is known as a correspondence.
We adopt a standard three-stage method:
\begin{enumerate} \item \textit{Detection}: Identify $K$ keypoints in each image. \item \textit{Description}: Generate keypoint descriptors as feature vectors in $\mathbb{R}^c$. \item \textit{Matching}: Descriptors are matched using mutual nearest neighbours. \end{enumerate}
This method encompasses both traditional techniques like SIFT \cite{lowe2004distinctive} and modern deep learning approaches. A notable example is DeDoDe~\cite{edstedt2024dedode}, which first optimizes keypoint detection from Structure from Motion reconstructions and then leverages the keypoint detector to train a descriptor through maximizing matching likelihood.
Our focus is on improving the second stage by explicitly handling affine transformations via \emph{steerers}, as introduced in \cite{bökman2024steerers}
for rotation equivariant keypoint description.
We generalize the formulation from rotations to locally affine transformations
in the remainder of this section, before explaining how to implement and train
affine steerers in the next section.

The general pipeline is outlined in Figure~\ref{fig:overview}.
We consider images as functions $I:\mathbb{R}^2\to \mathbb{R}^3$ and
general feature maps as functions $F:\mathbb{R}^2\to \mathbb{R}^c$.
Feature maps are obtained from feature extractors $f$, \ie, we will write $f(I)=F$.
Implicitly, we assume here that for a feature map $F$ associated to an image $I$,
the feature $F(x)$ at location $x$ is associated to the image content $I(x)$ at $x$.
The idea for keypoint description is that given a feature map $F$ and keypoints $x_i$, we obtain keypoint descriptions $d_i$ by evaluating $F$ at the locations $x_i$.
Images and feature maps can be warped by geometric image transformations
$\phi:\mathbb{R}^2 \to \mathbb{R}^2$,
by using the warp operator $W_\phi$ defined by
\begin{equation}
    W_\phi[I](x) = I(\phi^{-1} (x)).
\end{equation}
We will consider differentiable $\phi$,
so that they are locally approximated by affine transformations\footnote{In practice, image warps corresponding to camera motions are piecewise continuous and piecewise differentiable. The discontinuities stem from motion boundaries, which we will ignore in the theoretical part of this paper.}. \Ie,
for all $x\in\mathbb{R}^2$ there is an $M(x)\in\mathrm{GL}(2)$ s.t.\
\begin{equation}
    \phi(x+\Delta x) \approx \phi(x) + M(x) \Delta x,
\end{equation}
for small $\Delta x$.
$M:\mathbb{R}^2\to \mathrm{GL}(2)$ maps image locations
to local image transformations (affine warps) associated with $\phi$
at each point.
For some classes of image transformations $\phi$,
$M(x)$ takes a special form.
For instance, if $\phi\in\mathrm{SE}(2)$ is a roto-translation,
then $M(x)$ will always be a rotation matrix and so the possible $M(x)$'s
for roto-translations
form the subgroup $\mathrm{SO}(2)$ of $\mathrm{GL}(2)$.
Generally, given a set $\Phi$ of image transformations,
we can associate with it the minimal subgroup of $\mathrm{GL}(2)$
that contains all possible local image transformations $M(x)$.

Given a feature extractor $f$,
it is of interest how the extracted features
change when the image is warped by $W_\phi$.
For example, we could have
\begin{equation} \label{eq:equiv_trivial}
    f(W_\phi[I]) = W_\phi[f(I)],
\end{equation}
so that the same feature is extracted regardless of
whether $I$ is warped or not.
Introducing the notation $F_\phi$ for the feature map $f(W_\phi[I])$ extracted
from an image warped by $\phi$, we can evaluate \eqref{eq:equiv_trivial}
at $\phi(x)$ to get the form
\begin{equation} \label{eq:equiv_trivial_coord}
    F_\phi(\phi(x)) = F(x).
\end{equation}
So the feature at position $\phi(x)$ in $F_\phi$ is the same as the feature
at position $x$ in $F$.
When we have a collection of $\phi$'s forming a group
structure, $f$ satisfying \eqref{eq:equiv_trivial}
is said to to be equivariant.
This is however the least interesting form of equivariance where $f$ has to extract
the exact same feature for a given image content,
regardless of how the image is warped by $\phi$.
For example, a roto-translation-equivariant keypoint descriptor would describe all
rotations of a corner patch with the same description according to \eqref{eq:equiv_trivial}.
For this reason, descriptors satisfying \eqref{eq:equiv_trivial} for the group of rotations and translations are sometimes
also called invariant~\cite{bokman2022case}.
In \cite{bökman2024steerers} it was found that this leads to non-discriminative descriptions
on upright images, and the solution was to allow more general transformations of the features
using \emph{steerers}. Steerers are group representations, a concept from abstract algebra which we introduce next.

\begin{definition}[Group representation]
    Given a group $G$ and a vector space $V$, a group representation of $G$ on $V$ is a
    map $\rho: G\to \mathrm{GL}(V)$ such that
    \begin{equation}\label{eq:rep_condition}
        \rho(g_2 g_1) = \rho(g_2)\rho(g_1)\quad\text{for all $g_1, g_2\in G$.}
    \end{equation}
\end{definition}
Here, $\mathrm{GL}(V)$ is the general linear group of $V$, \ie, the group of all invertible linear maps $V \to V$ with composition as group operation.
For instance, we write $\mathrm{GL}(\mathbb{R}^n)=\mathrm{GL}(n)$ for
the group of invertible $n\times n$ matrices.
A group representation of $G$ on $\mathbb{R}^n$ is a map
from $G$ to invertible $n\times n$ matrices such that the group operation in $G$ corresponds
to matrix multiplication in $\mathrm{GL}(n)$.
Equipped with this formalism, we can now define steerers.

\begin{definition}[Steerer]\label{def:steerer}
    Let a set $\Phi$ of image transformations and the 
    corresponding group $G< \mathrm{GL}(2)$ of local affine transformations $M(x)$
    be given.
    For a feature extractor
    $f$ mapping images $I:\mathbb{R}^2 \to \mathbb{R}^3$ to
    feature maps $F:\mathbb{R}^2 \to \mathbb{R}^{c}$, a \emph{steerer} is a group representation 
    $\rho$ of $G$ on $\mathbb{R}^c$ such that $f$ becomes
    \emph{equivariant} w.r.t. $W_\phi$ and $\rho$, \ie,
    \begin{equation}\label{eq:equiv}
        f(W_\phi[I]) = \rho(W_\phi[M])W_\phi[f(I)].
    \end{equation}
\end{definition}    
    Using the notation $F_\phi$ for the feature map $f(W_\phi[I])$
    extracted from an image warped by $\phi$, we can evaluate \eqref{eq:equiv}
    at coordinate $\phi(x)$ as
    \begin{equation}\label{eq:equiv_coord}
        F_\phi(\phi(x)) = \rho(M(x))F(x).
    \end{equation}
    In words, the feature $F_\phi(\phi(x))$ describing the warped image at $\phi(x)$,
    is equal to the feature describing the original image at $x$,
    \ie $F(x)$, up to multiplication by
    the steerer $\rho(M(x))$.
    The steerer compensates for the local image transformation $M(x)$
    from coordinate $x$ to $\phi(x)$.
    We call $\rho$ a steerer even if \eqref{eq:equiv} only holds 
    approximately, as this will be the case in practice.

The reader should contrast \eqref{eq:equiv_trivial} and \eqref{eq:equiv_trivial_coord} with \eqref{eq:equiv} and \eqref{eq:equiv_coord}.
The more general \eqref{eq:equiv} and \eqref{eq:equiv_coord} allow features to change when the image content is warped,
but they must change in a predictible manner, specifically they must change by the
linear map $\rho(M)$, \ie, the steerer.
Our definition of steerers generalizes the one in \cite{bökman2024steerers},
from $\Phi$ only containing roto-translations, to arbitrary differentiable
image transformations.
Utilizing steerers, Bökman \etal~\cite{bökman2024steerers} successfully trained rotation equivariant keypoint descriptors that match or surpass the performance of non-equivariant keypoint descriptors on upright images, while also facilitating rotation invariant matching.
We will only be concerned with the case $G=\mathrm{GL}(2)$ in this paper.
Specifically, having access to a steerer means that one can compute what the features
$f(W_\phi[I])$ for a warped image according to any transformation $\phi$ 
would be, given only the features $f(I)$ from the original image.
This enables test time augmentation in feature space, \ie, computing features
for multiple warps $W_\phi$ while only running the feature extractor $f$ once.

\section{Method}\label{sec:method}
We outline the relevant mathematical theory for $\mathrm{GL}(2)$-steerers in Section~\ref{sec:gl2}, followed by an a delineation from rotation steerers in Section~\ref{sec:rot_steer_diff} and an explanation of their integration with the keypoint descriptor neural network in Section~\ref{sec:matcher}. Subsequent sections, from Section~\ref{sec:train-megadepth} to Section~\ref{sec:finetune}, detail the comprehensive training steps for the entire pipeline.

\subsection{Representation theory of \texorpdfstring{$\mathrm{GL}(2)$}{GL(2)} \label{sec:gl2}}
In this paper, keypoint descriptions will be vectors in 
$\mathbb{R}^{256}$, which we will call description space.
The aim of this section is to explain how we can build a steerer $\rho$
(Definition~\ref{def:steerer})
for $\mathrm{GL}(2)$ on description space.
From now we will write elements of $\mathrm{GL}(2)$ as matrices
\begin{equation}
M = \begin{pmatrix}
\alpha & \beta \\
\gamma & \delta
\end{pmatrix}, \quad \text{s.t.~} \alpha\delta - \beta\gamma \neq 0.
\end{equation}
Understanding what representations $\rho$ for $\mathrm{GL}(2)$
can look like is a well studied problem in mathematical representation theory,
presented for instance in \cite[Chapter 4]{olverClassicalInvariantTheory1999}.
We refer to \cite{olverClassicalInvariantTheory1999} for details and will here limit ourselves to 
explaining the way of constructing representations $\rho$ that we will use
in the experiments.

\begin{example}[Trivial representation]
    The simplest form of $\rho$ is the constant map $\rho(M)=1$,
    which trivially satisfies \eqref{eq:rep_condition} and is a one-dimensional representation.
\end{example}

\begin{example}[Determinant representation]
    Slightly more involved, we can construct representations using
    the determinant of $M$.
    For instance,
    \begin{equation}\label{eq:det_rep_int}
        \rho(M) = 
            \mathrm{det}(M)^{n} \quad\mbox{and}\quad
        \rho(M) = 
            |\mathrm{det}(M)|^{\xi}
    \end{equation}
    satisfy \eqref{eq:rep_condition} for any choice of 
    $n\in \mathbb{N}$ and  $\xi\in \mathbb{R}$.
\end{example}

\begin{example}[Standard representation]
    A straightforward way to construct a representation of
    $\mathrm{GL}(2)$ on $\mathbb{R}^{2}$ is to take the
    matrix itself: 
        $\rho(M) = M$.
\end{example}

The easiest way to construct representations of larger dimensions is to stack
smaller representations into block-diagonals.
We denote by $X\oplus Y$ the block diagonal matrix with blocks $X$ and $Y$.
So, for instance, $\rho(M)=\bigoplus_{i=1}^{128} M$ is a representation on $\mathbb{R}^{256}$.
It is also easy to see that any change of basis of a representation defines a new representation: $\tilde\rho(M)=Q^{-1}\rho(M)Q$, where $Q$ is some invertible matrix.
In our experiments we want to use a broad range of representations,
and the idea will be to build them out of so called irreducible representations, irreps\footnote{An irrep on $V$ is a representation that does not leave any proper subspace $W\subset V$ invariant.
Irreps can be thought of as fundamental building blocks of representations as many general representations can be decomposed into irreps. However, for $\mathrm{GL}(2)$, not all representations can be built out of its irreps. The standard counterexample is $\rho(M)=\begin{smallpmatrix}1 & \log|\det M| \\ 0 & 1\end{smallpmatrix}$~\cite[Example 4.11]{olverClassicalInvariantTheory1999}.}.
Given a set of irreps $\rho_i$ we will form larger representations on description space by $\rho(M) = Q^{-1}(\bigoplus_{i=1}^n\rho_i(M))Q$.

To find irreps of $\mathrm{GL}(2)$, one can look at vector spaces
of homogeneous polynomials in two variables.
The following example demonstrates the general principle.
\begin{example}[Homogeneous quadratics]\label{ex:rep_quad}
    Consider the vector space $\mathcal{H}_2$ of
    homogeneous quadratic polynomials in two variables $x, y$.
    It has a basis consisting of the monomials $y^2, 2xy, x^2$ and
    so any $q\in\mathcal{H}_2$ can be written
    \begin{equation}
        q(x, y) = a_0 y^2 + a_1 2xy + a_2 x^2
    \end{equation}
    with some coefficients $a_0, a_1, a_2\in\mathbb{R}$.
    The reader will note that $\mathcal{H}_2$ is
    isomorphic to $\mathbb{R}^3$,
    since polynomials $q$ are in bijection with triples
    $(a_0, a_1, a_2)$.
    The reason to think about these
    triples as polynomials is that we can 
    in a natural way find a representation of
    $\mathrm{GL}(2)$ on $\mathcal{H}_2$.
    We define the representation $\rho_2$ on $\mathcal{H}_2$ by
    \begin{equation}\label{eq:rep_fun}
        (\rho_2(M) q)(x, y) = q((x, y) M),
    \end{equation}
    where $(x, y) M$ denotes multiplying the row-vector $(x, y)$
    by the matrix $M$.
    This satisfies \eqref{eq:rep_condition} as
    \begin{equation}
    \begin{split}
        (\rho_2(M_2 M_1) q)(x, y)
        &= q((x, y)(M_2 M_1)) = q(((x, y)M_2)M_1) 
        \\ &= (\rho_2(M_1)q)((x, y)M_2) = (\rho_2(M_2)(\rho_2(M_1)q))(x, y).
    \end{split}
    \end{equation}
    We can rewrite $\rho_2$ as a representation on $\mathbb{R}^3$
    by considering how the coefficients of $q$ change under \eqref{eq:rep_fun}:
    \begin{equation}
    \begin{split}
        (\rho_2(M)q)(x,y) 
        &= q((x, y)M) = q(\alpha x + \gamma y, \beta x + \delta y) 
        \\ &= a_0 (\beta x + \delta y)^2 
        + a_1 2(\alpha x + \gamma y)(\beta x + \delta y)
        + a_2 (\alpha x + \gamma y)^2
        \\ &= \left( \begin{pmatrix}
            \delta^2 & 2\gamma\delta & \gamma^2 \\
            \beta\delta & \alpha\delta + \beta\gamma & \alpha\gamma \\
            \beta^2 & 2\alpha\beta & \alpha^2
        \end{pmatrix} \begin{pmatrix} a_0 \\ a_1 \\ a_2 \end{pmatrix} \right)
        \cdot \begin{pmatrix} y^2 \\ 2xy \\ x^2 \end{pmatrix}.
    \end{split}
    \end{equation}
    So we have found the action of $\rho_2$ on the polynomial coefficients $(a_0, a_1, a_2)^T\in\mathbb{R}^3$ and can by slight abuse of notation
    write $\rho_2(M)$ on matrix form as
    \begin{equation}\label{eq:rep_poly_matrix}
        \rho_2(M) =
        \begin{pmatrix}
            \delta^2 & 2\gamma\delta & \gamma^2 \\
            \beta\delta & \alpha\delta + \beta\gamma & \alpha\gamma \\
            \beta^2 & 2\alpha\beta & \alpha^2
        \end{pmatrix}.
    \end{equation}
\end{example}

Constructing higher dimensional irreps 
can be done by following the recipe outlined in
Example~\ref{ex:rep_quad}.
The $(n+1)$-dimensional vector space of homogeneous polynomials of degree $n$
\begin{equation}
    \mathcal{H}_n = \mathrm{span}\left\{\begin{pmatrix}n \\ k\end{pmatrix} x^k y^{n-k}\right\}_{k=0}^n
\end{equation}
has an associated representation $\rho_n$ of $\mathrm{GL}(2)$ defined by \eqref{eq:rep_fun}.
$\rho_n(M)$ can be written on matrix form similar to \eqref{eq:rep_poly_matrix},
where each entry of the matrix is a homogeneous degree $n$ polynomial in
$\alpha, \beta, \gamma, \delta$.
A formula for $\rho_n(M)$ is given by \cite[(2.6)]{olverClassicalInvariantTheory1999}.

More general representations can be obtained by
multiplying with determinant based representations and we define
\begin{equation}\label{eq:rep_general}
    \rho_{n,\xi}(M) = |\mathrm{det}(M)|^{\xi-\frac{n}{2}} \rho_n(M).
\end{equation}
where the power $-\frac{n}{2}$ is introduced to normalize the determinant of $\rho_{n,0}$ to $\pm 1$. These $\rho_{n,\xi}$ are irreps and form the basic building blocks of our steerers.
The steerers we use in the experiments are formed as
\begin{equation}\label{eq:rep_desc}
    \rho(M) = Q^{-1} \left(\bigoplus_j \rho_{n_j, \xi_j}(M)\right) Q,
\end{equation}
where $Q\in \mathbb{R}^{256\times 256}$ is a learnt change of basis matrix initialized to identity,
$\xi_j$ are learnt determinant scalings and
$n_j\in\{0, 1, 2, 3, 4\}$ are chosen to get an %
equal distribution of dimensions of the different orders $0, \ldots, 4$.
The cutoff at $4$ is motivated by the fact that it covers rotation
frequencies up to $4$
when a rotation $R$ is fed into $\rho$, and higher frequencies were not useful in prior work\cite{bökman2024steerers, felsberg2002}.

\subsection{Comparison to rotation steerers} \label{sec:rot_steer_diff}
If we input a rotation matrix $R$ into \eqref{eq:rep_desc}, we obtain a steerer like those considered in \cite{bökman2024steerers}. In that case, $\det(R)=1$ so the determinant scaling does not affect the steerer.
The main novelty in this work is that we can now steer the much larger class of local transformations $\mathrm{GL}(2)$.
\eqref{eq:rep_desc} specifies a structured way for the descriptor to respond to affine transformations, where
the determinant scaling captures how much the descriptor varies with the scale of the transform $M$.

\begin{figure}[t]
    \centering
    \includegraphics[width=\textwidth,trim=0cm 5mm 0cm 5mm,clip=true]{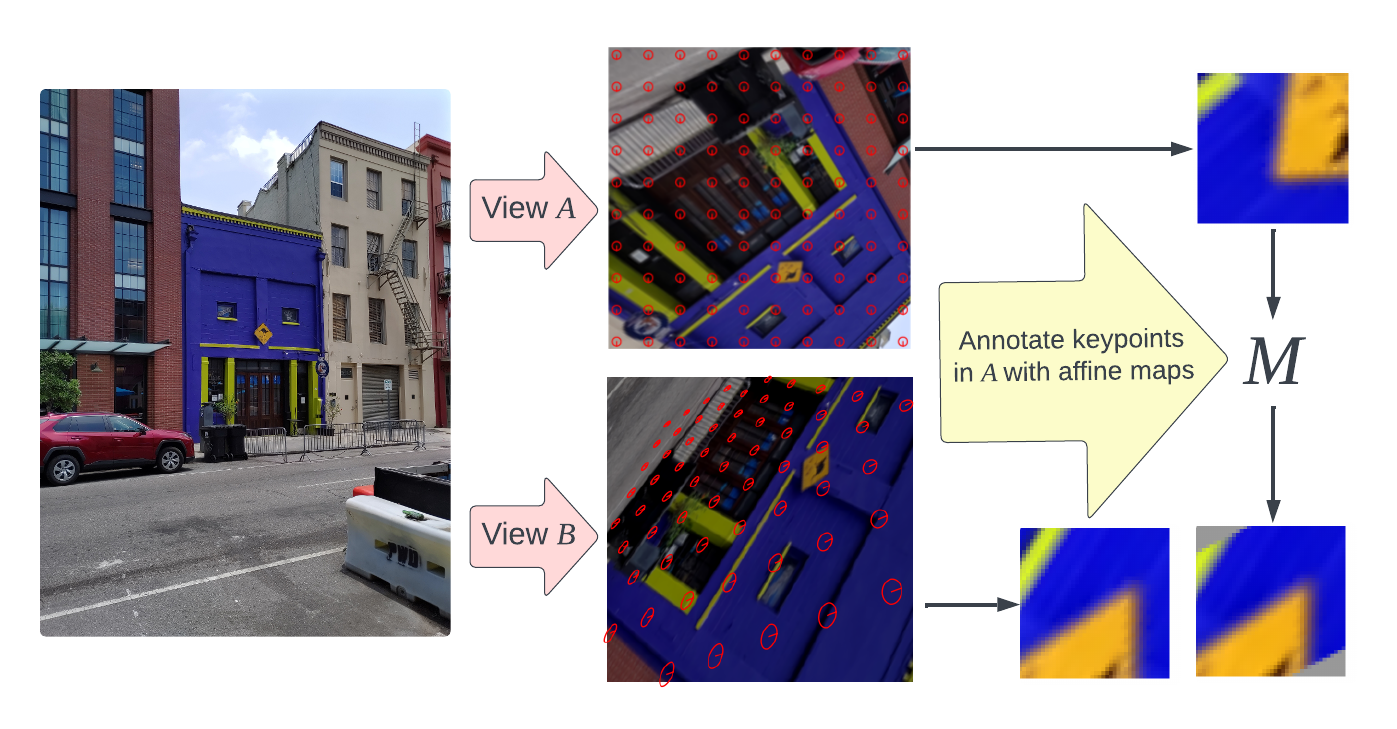}
    \caption{
    \textbf{Illustration of the data generation pipeline.} 
    Given two views $A$, $B$ of a scene, we compute the image warp from one to the other and use it to both find corresponding keypoints and annotate the keypoints in $A$ with affine transformations $M$ that locally approximate the image warp from $A$ to $B$.
    In the illustration above, we show two views obtained as homographies of a single image, as used in the pretraining step described in Section~\ref{sec:pretrain}, but it is also possible to obtain the two views, \eg, by taking two photos of the same location as is the case when we train on MegaDepth (Section~\ref{sec:train-megadepth}).
    The red circles in $A$ are warped to the ellipses in $B$, and we illustrate the obtained
    affine map $M$ for one keypoint pair on the right hand side of the figure.
    In practice we use DeDoDe keypoints but here we illustrate with a regular grid in $A$.
    }
    \label{fig:homography_data}
\end{figure}
\subsection{Max similarity matcher revisited \label{sec:matcher}}
Given two descriptions $d_1, d_2\in \mathbb{R}^{256}$,  
we want to be able to
say how close they are to each other in order to determine
correspondences between keypoints in different images.
The similarity measure used in \cite{edstedt2024dedode, bökman2024steerers} %
is the cosine similarity.
This similarity is however unsuited for affinely steered
descriptions, as the steerer \eqref{eq:rep_desc}
can change the norm of a description.
We therefore use Euclidean norm instead and measure
the similarity of $d_1$ and $d_2$ by
    $-\| d_1 - d_2 \|.$
In practice, given two images $A$ and $B$,
we find $K$ keypoints in each image,
describe them with descriptions
and measure all pairwise similarities $S\in
\mathbb{R}^{K\times K}$.
We get matches from the pairwise similarity matrix $S$
by taking mutual nearest neighbours and putting a minimum
threshold on the dual softmax score of $S$.

The similarity of descriptions can be more robustly
computed when a steerer is available.
In \cite{bökman2024steerers}, several methods were suggested,
one of which is the \emph{max similarity} method.
The similarity between $d_1$ and $d_2$ is then
\begin{equation}\label{eq:max_similarity}
    \max_{M\in \mathcal{M}}-\| \rho(M)d_1 - d_2 \|,
\end{equation}
where $\mathcal{M}$ is a set of transformations
that can be selected to balance
robustness over a range of transformations with inference
speed and discriminability.
We propose to learn $\mathcal{M}$ by backpropagating through \eqref{eq:max_similarity} during finetuning (Section~\ref{sec:finetune}), although this
limits us to a small set $\mathcal{M}$ for computational reasons.

\subsection{Training the descriptor and steerer on MegaDepth}\label{sec:train-megadepth}
We follow \cite{bökman2024steerers} and train the descriptor network $f$
and steerer $\rho$ \eqref{eq:rep_desc} jointly.
We train on MegaDepth \cite{li2018megadepth} similar to 
\cite{edstedt2024dedode, bökman2024steerers}, but
in contrast to them use heavy affine augmentations of the images.
Given an image pair $A, B$, we detect $K$ DeDoDe keypoints \cite{edstedt2024dedode}
$x_{A, i}$ ($i=1,\ldots,K$), resp. $x_{B, j}$ ($j=1,\ldots,K$) per image,
and obtain descriptions $d_{A, i}$, resp. $d_{B, j}$ by evaluating the extracted feature maps $f(A), f(B)$ at the keypoint locations.
Ground truth matches between keypoints in the two images are
obtained as in \cite{edstedt2024dedode}: by using the known image depths
and relative pose we warp the keypoints from $A$ to $B$
and from $B$ to $A$ and take matches as all mutual closest neighbours with distances
below $0.5\%$ of the image width.
For the ground truth matches, we also compute local affine maps $M_i$ from $A$ to $B$ by warping
a small octagon of points around each keypoint in $A$ to $B$ and taking
the least squares estimated affine map from the eight points in $A$ to
the eight points in $B$.
The local affine maps $M_i$ are used to steer the descriptions in $A$
during training in order to align them with the ones in $B$.
We also estimate a global affine $M$ by taking the least squares estimate
of all ground truth matches. This $M$ is used to steer all descriptions in $A$
which do not have a match in $B$.
In summary, during training, the similarity between a description $d_{A, i}$ in $A$
and one $d_{B, j}$ in $B$ is 
\begin{equation}\label{eq:sim_AB}
    S_{ij} = -\| \rho(M_i)d_{A, i} - d_{B, j} \|.
\end{equation}
We form the similarity matrix $S\in \mathbb{R}^{K\times K}$ by considering all
pairs of descriptions in $A$ and $B$ and compute the loss as in \cite{edstedt2024dedode} by taking
the negative dual log-softmax of $S$ and taking the mean over the ground truth matches.

\subsection{Pretraining on homographies}\label{sec:pretrain}
Training or pretraining descriptors and matchers on homographies
is a common strategy \cite{detone2018superpoint, sarlin2020superglue, gleize2023silk, lindenberger2023lightglue}.
The idea is to take a single image and match it with a homography warped
(and photometrically augmented) version of itself.
Homography pretraining is a good fit for affine steering, since
a homography applied to an image
induces an affine transform between each point in the
original image and the corresponding point in the warped image.
We visualize the data generation pipeline in Figure~\ref{fig:homography_data}.

We find that pretraining on homographies is not very helpful for
affine steerers trained by augmentations of MegaDepth.
However, pretraining on homographies gives a good starting point
for the fine-tuning approach explained next.

\subsection{Fine-tuning on upright images}\label{sec:finetune}
We propose a way to fine-tune a descriptor and steerer jointly to obtain
descriptions tailored for upright images.
We train through the \emph{max similarity} score~\eqref{eq:max_similarity},
where we use a set of only three affine transformations
$\mathcal{M} = \{I, M_1, M_2\}$, letting $M_1, M_2$ be learnable
``prototypes''.
The aim is that the learnable $M_1, M_2$ will contain affine transformations such
that the net does not have to become invariant under $M_1, M_2$ while still
enabling matching that is invariant under $M_1, M_2$ through \eqref{eq:max_similarity}.
Note that in contrast to the earlier described training approaches with the loss computed from \eqref{eq:sim_AB} we do not enforce a connection between $M_1, M_2$ and affine
warps of the image at this stage.
Hence there are no guarantees that steering by an $M\in\mathrm{GL}(2)$
will correspond to warping the input image by $M$ any longer.
For instance, steering by $M$ could correspond to warping the input image
by $\tilde M M \tilde M^{-1}$ for some $\tilde M$,
which would mean that the steerer that actually
corresponds to warping by $M$ is $\tilde\rho(M) = \rho(\tilde M)^{-1}\rho(M)\rho(\tilde M)$.
Even more generally, the steerer could learn to incorporate other image
transformations than affine warps into the learnt prototypes $M_1, M_2$.
We show numerically how well the steerer works for affine warps after fine-tuning
in Section~\ref{sec:steer_cap}.

\begin{table}[t]
\parbox{.45\linewidth}{
 
     \centering
     \caption{\textbf{Megadepth-1500}. Relative pose estimation, measured in AUC (higher is better).}
     \begin{tabular}{l rrr}
     \toprule
      Method $\downarrow$\quad\quad\quad AUC $\rightarrow$& $@5^{\circ}$&$@10^{\circ}$&$@20^{\circ}$\\
\midrule
SuperPoint~\cite{detone2018superpoint}~\tiny{CVPRW'18} & 31.7 & 46.8 & 60.1 \\
DISK~\cite{tyszkiewicz2020disk}~\tiny{NeurIPS'20} & 36.7 & 52.9 & 65.9 \\
ALIKED~\cite{Zhao2023ALIKED}~\tiny{IEEE-TIM'23} & 41.9 & 58.4 & 71.7 \\
SiLK~\cite{gleize2023silk}~\tiny{ICCV'23} & 39.9 & 55.1 & 66.9 \\
\midrule
DeDoDe-B~\cite{edstedt2024dedode}~\tiny{3DV'24} & 49.4 & 65.5 & 77.7 \\
Steerers-B-C4-Perm~\cite{bökman2024steerers}& 51\phantom{.0} & 67\phantom{.0} & 79\phantom{.0} \\
\emph{AffEqui-B} & 46.1 & 62.4 & 74.8 \\
\emph{AffSteer-B} & \textbf{52.7} & \textbf{68.9} & \textbf{81.0} \\
\midrule
DeDoDe-G~\cite{edstedt2024dedode}~\tiny{3DV'24} & 52.8 & 69.7 & 82.0 \\
Steerers-G-C4-Perm~\cite{bökman2024steerers}& 52\phantom{.0} & 69\phantom{.0} & 81\phantom{.0} \\
\emph{AffEqui-G} & 50.6 & 67.2 & 80.1 \\
\emph{AffSteer-G} & \textbf{53.7} & \textbf{70.0} & \textbf{82.1} \\

     \bottomrule
     \end{tabular}
     \label{tab:megadepth-loftr}

}
\hfill
\parbox[c]{.45\linewidth}{

\centering
\caption{\textbf{WxBS~\cite{mishkin2015WXBS}.} Fundamental matrix estimation, measured in mAA at 10px (higher is better). } 

\begin{tabular}{ l  l  }
  \toprule
        Method $\downarrow$\quad\quad\quad mAA$@$ $\rightarrow$& $10\text{px}$ \\
 \midrule
DISK~\cite{tyszkiewicz2020disk}~\tiny{NeurIPS'20}& 35.5\\
\midrule
DeDoDe-B~\cite{edstedt2024dedode}~{\tiny{3DV'24}} & $45.9\pm 1.2$\\
Steerers-B-C4-Perm~\cite{bökman2024steerers}& $\mathbf{51.0}\pm 1.5$\\
\emph{AffSteer-B} & $50.0\pm 1.4$\\
\midrule
DeDoDe-G~\cite{edstedt2024dedode}~{\tiny{3DV'24}} & $\mathbf{57.7}\pm 1.1$\\
Steerers-G-C4-Perm~\cite{bökman2024steerers} & $57.0\pm 1.4$\\
\emph{AffSteer-G} & $57.3\pm 1.2$\\
  \bottomrule
\end{tabular}
\label{tab:WxBS}

}
\end{table}

\section{Experiments}\label{sec:experiments}
Experiment details and ablations can be found in the appendix.
We follow \cite{edstedt2024dedode} and consider two descriptor networks, a \emph{B}-model with a VGG-19 backbone \cite{simonyan2015vgg} and a larger \emph{G}-model with a DINOv2 ViT backbone\cite{oquab2023dinov2, dosovitskiy2021vit}.
In the appendix we also present an experiment with the smaller XFeat descriptor~\cite{potje2024xfeat}.

For both \emph{B}- and \emph{G}-networks, we have two versions.
One version is optimized for upright images by pretraining on homographies (Section~\ref{sec:pretrain}) and fine-tuning on upright MegaDepth pairs (Section~\ref{sec:finetune}).
These models are called \emph{AffSteer-B} and \emph{AffSteer-G}.
During inference, we use the set of three affine transformations $\mathcal{M}$ learnt during training for the \emph{max similarity} matching method \eqref{eq:max_similarity}.

The second version is trained on Megadepth pairs that are affinely augmented (Section~\ref{sec:train-megadepth}).
This version is not as good on upright images, but can be steered with
affine warps and is more equivariant.
These models are called \emph{AffEqui-B} and \emph{AffEqui-G}.
We compare the equivariance and steering capabilities of the versions in Section~\ref{sec:steer_cap}.

We evaluate on several common benchmarks for upright image matching.
Here we consistently set new state-of-the-art results for
detector-descriptor based methods.
Dense methods such as the SotA~\cite{edstedt2024roma} are heavier
and take two images jointly as input, allowing them
to compute features in each image conditioned on the other image.
Descriptor based methods cannot do this and so we don't compare
to dense methods here.
Generally, we divide the tables into three sections of 
first works not based on DeDoDe~\cite{edstedt2024dedode}, then
methods with the same network architecture as DeDoDe-B and
finally methods with the same network architecture as DeDoDe-G.

\paragraph{MegaDepth-1500 relative pose.}
We follow LoFTR~\cite{sun2021loftr} and evaluate on a held out part of MegaDepth.
This is a standard benchmark for matching upright images.
We present results in Table~\ref{tab:megadepth-loftr}, where we see consistent improvements for the upright optimized \emph{AffSteer}-models, while the \emph{AffEqui}-models perform competitively with previous approaches.%

\begin{table}[t]
\parbox{.4\linewidth}{

    \centering
    \caption{\textbf{%
    IMC2022~\cite{image-matching-challenge-2022}.} %
    Relative pose %
    on
    the hidden test set,
    measured in mAA (higher is better). }    \label{tab:imc2022}
    \begin{tabular}{ll}
    \toprule
     Method $\downarrow$\quad\quad\quad mAA $\rightarrow$&$@10$\\
     \midrule

DISK~\cite{tyszkiewicz2020disk}~\tiny{NeurIPS'20} & 64.8 \\
ALIKED~\cite{Zhao2023ALIKED}~\tiny{IEEE-TIM'23} & 64.9 \\
SiLK~\cite{gleize2023silk}~\tiny{ICCV'23} & 68.5 \\
\midrule
DeDoDe-B~\cite{edstedt2024dedode}~\tiny{3DV'24} & 72.9 \\
Steerers-B-C4-Perm~\cite{bökman2024steerers}~\tiny{CVPR'24} & 73.4 \\
\emph{AffSteer-B} &  \textbf{77.3}\\
\midrule
DeDoDe-G~\cite{edstedt2024dedode}~\tiny{3DV'24} & 75.8 \\
Steerers-G-C4-Perm~\cite{bökman2024steerers}~\tiny{CVPR'24} & 75.5 \\
\emph{AffSteer-G} &  \textbf{76.8}\\
\bottomrule
    \end{tabular}

}
\quad\enspace
\parbox{.5\linewidth}{
    \centering
    \caption{\textbf{HPatches~\cite{balntas2017hpatches} Homography.} Corner error, measured in AUC (higher is better).}%
    \begin{tabular}{lrrr}
    \toprule
     Method $\downarrow$\quad\quad\quad AUC$@$ $\rightarrow$       & $3$px & $5$px & $10$px\\
       \midrule
    DISK~\cite{tyszkiewicz2020disk}~\tiny{NeurIPS'20} & 60.3 & 71.4 & 81.8 \\
    ALIKED~\cite{Zhao2023ALIKED}~\tiny{IEEE-TIM'23} & 61.6 & 73.1 & 83.5\\
    SiLK~\cite{gleize2023silk}~\tiny{ICCV'23} & 66\phantom{.0} & - & - \\
    \midrule
    DeDoDe-B~\cite{edstedt2024dedode}~\tiny{3DV'24} & 68.2 & 77.9 & 86.4 \\
    Steerers-B-C4-Perm~\cite{bökman2024steerers}~\tiny{CVPR'24} & 69.5 & 78.7 & 87.0 \\
    \emph{AffSteer-B} & \textbf{70.1} & \textbf{79.1} & \textbf{87.3} \\
    \midrule
    DeDoDe-G~\cite{edstedt2024dedode}~\tiny{3DV'24} & 67.1 & 77.3 & 86.3 \\
    Steerers-G-C4-Perm~\cite{bökman2024steerers}~\tiny{CVPR'24} & 66.8 & 76.7 & 85.9\\

    \emph{AffSteer-G} & \textbf{68.2} & \textbf{78.1} & \textbf{86.9} \\
    \bottomrule
    \end{tabular}
    \label{tab:hpatches_homog}

}    
\end{table}

\paragraph{WxBS Fundamental matrix estimation.}
Results for the WxBS benchmark~\cite{mishkin2015WXBS} and are shown in Table~\ref{tab:hpatches_homog}.
Here we find that the results fluctuate a lot due to RANSAC and
dual-soft-max thresholds. 
We report results with the best found thresholds
for each method and mean and standard deviation over 10 evaluations.

\begin{table}[t]
\small
\centering
\caption{\textbf{AIMS~\cite{stoken2023astronaut}.} Measured in average precision (higher is better). The subset ``North Up'', contains image pairs
with small relative rotation and ``All Others'' contains image pairs with large relative rotations. ``All'' includes all image pairs.} 

\begin{tabular}{ l l r  r r }
  \toprule
        Matching method~~~~ & Descriptor & North Up &\quad All Others &\quad All \\
 \midrule
Max similarity 8 & Steerers-B-SO2-Spread~\cite{bökman2024steerers}& 60 & 57 & 58\\
Procrustes & Steerers-B-SO2-Freq1~\cite{bökman2024steerers}& \textbf{64} & \textbf{59} & \textbf{60}\\
Max similarity 8 & \emph{AffEqui-B} & 59 & 58 & 59 \\
Max similarity 12 &\emph{AffEqui-B} & 61 & \textbf{59} & \textbf{60} \\
  \bottomrule
\end{tabular}
\label{tab:aims}
\end{table}
\paragraph{Image Matching Challenge 2022.}
Results for the Image Matching Challenge 2022~\cite{image-matching-challenge-2022} are presented in Table~\ref{tab:imc2022}.
Remarkably, we obtain a large improvement for \emph{B}-size
variants,
even outperforming all \emph{G}-size variants with \emph{AffSteer-B}.
As the test set is hidden, we unfortunately
cannot offer an explanation as to what types of
image pairs we are handling better than previous works.

\paragraph{HPatches Homography Estimation.}
We evaluate on the HPatches Homography benchmark~\cite{balntas2017hpatches,sun2021loftr}. We present results in Table~\ref{tab:hpatches_homog}. 
Interestingly, \emph{B}-size DeDoDe variants do better on homographies than \emph{G}-size methods generally.

\paragraph{Astronaut Image Matching Challenge (AIMS).}
To check how well our equivariant models handle rotations,
we compare against rotation steerers~\cite{bökman2024steerers} on AIMS~\cite{stoken2023astronaut}.
Here we test a \emph{max similarity} matcher with $8$ or $12$ rotations and
use the \emph{AffEqui-B} model, since~\cite{bökman2024steerers} also use 
a \emph{B}-model in their evaluation.
We show competitive performance even though
our method is not specialized for rotations.
In particular, we cannot use the Procrustes matcher proposed in \cite{bökman2024steerers}, but show that we can match its performance
by increasing the amount of rotations in the max similarity matcher.

\begin{figure}[t]
    \centering
\begin{tikzpicture}
    \small
  \centering
  \begin{semilogyaxis}[
        ybar, axis on top,
        title={},
        height=.21\textwidth, width=\textwidth,
        bar width=0.35cm,
        ymajorgrids, tick align=inside,
        major grid style={draw=white},
        enlarge y limits={value=.1,upper},
        ymin=100, ymax=8000,
        axis x line*=bottom,
        axis y line*=right,
        y axis line style={opacity=0},
        tickwidth=0pt,
        enlarge x limits=true,
        legend style={
            at={(0.5,-0.45)},
            anchor=north,
            legend columns=-1,
            /tikz/every even column/.append style={column sep=0.5cm}
        },
        ylabel={Correct matches},
        symbolic x coords={
           No affine, Scale 1 + Rot, Scale 2 + Rot, Scale 2 + Rot + Norm, },
       xtick=data,
       xticklabels={
           No affine, Scale 1 + Rot, Scale 2 + Rot, Scale 2$^*$ + Rot \\, },
       xticklabel style={align=center},
    ]
    \addplot [draw=none, fill=red!30] coordinates {
      (No affine,7457)
      (Scale 1 + Rot,6962) 
      (Scale 2 + Rot,5675)
      (Scale 2 + Rot + Norm,5443) 
    };
    \addplot [draw=none, fill=blue!30] coordinates {
      (No affine,7613)
      (Scale 1 + Rot,7016) 
      (Scale 2 + Rot,5626)
      (Scale 2 + Rot + Norm,5516) 
    };
    \addplot [draw=none, fill=green!50] coordinates {
      (No affine,7618)
      (Scale 1 + Rot,789) 
      (Scale 2 + Rot,320)
      (Scale 2 + Rot + Norm,649) 
    };
    \addplot [draw=none, fill=yellow!90] coordinates {
      (No affine,7680)
      (Scale 1 + Rot,2194) 
      (Scale 2 + Rot,1328)
      (Scale 2 + Rot + Norm,1784) 
    };

    \legend{\emph{AffEqui-B},\emph{AffEqui-G},\emph{AffSteer-B}, \emph{AffSteer-G}}
  \end{semilogyaxis}
  \end{tikzpicture}
\caption{
    Measuring the affine steering capabilities of the different descriptors.
    We count correct matches when using the steerer with the ground truth affine
    map on affinely augmented images. We vary the affine maps by scale.
    ``+ Rot'' means that all inplane rotations are included.
    ``Scale 2$^*$'' means that the scale
    of the affine map $M$ is maximum 2, but the matrix that is fed to the
    steerer is $M/\sqrt{\det{M}}$ and so has determinant 1.
}
\label{fig:oracle-steer}
\end{figure}
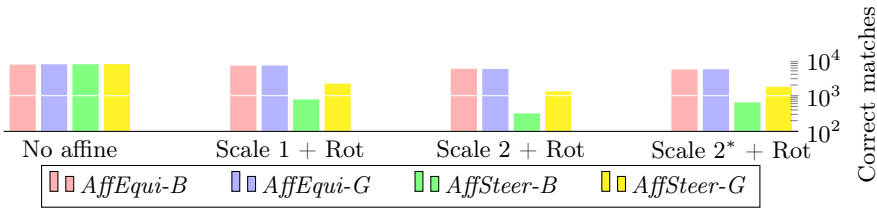
\subsection{Evaluating the steering capabilities}\label{sec:steer_cap}
To evaluate how well the steerer works for affine steering,
we take a subset of MegaDepth images and investigate how
many matches are obtained when matching with an \emph{oracle steerer} to a
affinely augmented version of each image.
In other words, we measure description similarity by
\begin{equation}
    -\|\rho(M_i)d_{A,i} - d_{B,j}\|,
\end{equation}
where $M_i$ is the correct affine warp.
We will measure the number of correct matches across various difficulties of affine warps.
If the steerer works well, then we should get similarly many matches
in the warped cases as in the case where there is no warp.
We show the results in Figure~\ref{fig:oracle-steer}.
It is clear that the steerer works well for 
\emph{AffEqui} as expected.
To further confirm this, we evaluate \emph{AffEqui} on HPatches with an oracle steerer that steers the descriptions with the correct $M_i\in\mathrm{GL}(2)$.
The results are (71.2, 80.8, 89.2) AUC@(3px, 5px, 10px) for \emph{AffEqui-B} and
(70.9, 80.1, 88.7) for \emph{AffEqui-G}.
These numbers are comparable to using the SotA dense matcher RoMa~\cite{edstedt2024roma}, which achieves (71.3, 80.6, 88.5).
While the comparison is unfair in the sense that we have access to the correct $M_i$,
this still indicates a large potential of affinely equivariant descriptors.
We hope that it will be possible in the future to leverage this potential.

As seen in Figure~\ref{fig:oracle-steer}, for \emph{AffSteer-G} the steerer still works quite well, but
the steering of scale seems to have been destroyed by the
fine-tuning, since we get more matches when normalizing the correct affine map $M$ to unit determinant before feeding it into the steerer.
For \emph{AffSteer-B} the steerer does not work well.
A possible explanation is given in Section~\ref{sec:finetune},
the steerer $\rho(M)$ might no longer directly encode an affine image warp by $M$ but some other transformation.

\section{Conclusion}\label{sec:conclusion}
We presented a generalization of the steerers framework to locally affine transformations.
This led to affinely equivariant descriptors \emph{AffEqui-B},
\emph{AffEqui-G}.
These produced results below the state-of-the-art, but showed promise for combining with estimation of local affine transforms, which we explored through the use of an oracle method.
Then we proposed a fine-tuning method for upright images, producing the state-of-the-art descriptors \emph{AffSteer-B} and \emph{AffSteer-G}.

\newpage
\section*{Acknowledgements}
This work was supported by the Wallenberg Artificial
Intelligence, Autonomous Systems and Software Program
(WASP), funded by the Knut and Alice Wallenberg Foundation and by the strategic research environment ELLIIT, funded by the Swedish government. The computational resources were provided by the
National Academic Infrastructure for Supercomputing in
Sweden (NAISS) at C3SE, partially funded by the Swedish Research
Council through grant agreement no.~2022-06725, and by
the Berzelius resource, provided by the Knut and Alice Wallenberg Foundation at the National Supercomputer Centre.

\bibliographystyle{splncs04}
\bibliography{main}

\newpage
\appendix
\begin{table}[h]
     \centering
     \caption{\textbf{Ablations on Megadepth-1500}. Relative pose estimation, measured in AUC (higher is better). The last section of the table contain ablations described in Section~\ref{sec:ablation}.}
     \begin{tabular}{l rrr}
     \toprule
      Method $\downarrow$\qquad\qquad\qquad\qquad\qquad\qquad\qquad AUC $\rightarrow$& $@5^{\circ}$&$@10^{\circ}$&$@20^{\circ}$\\
\midrule
DeDoDe-B~\cite{edstedt2024dedode} & 49.4 & 65.5 & 77.7 \\
Steerers-B-C4-Perm~\cite{bökman2024steerers} & 51\phantom{.0} & 67\phantom{.0} & 79\phantom{.0} \\
\emph{AffEqui-B} & 46.1 & 62.4 & 74.8 \\
\emph{AffSteer-B} & \textbf{52.7} & \textbf{68.9} & \textbf{81.0} \\
\midrule
DeDoDe-G~\cite{edstedt2024dedode} & 52.8 & 69.7 & 82.0 \\
Steerers-G-C4-Perm~\cite{bökman2024steerers} & 52\phantom{.0} & 69\phantom{.0} & 81\phantom{.0} \\
\emph{AffEqui-G} & 50.6 & 67.2 & 80.1 \\
\emph{AffSteer-G} & \textbf{53.7} & \textbf{70.0} & \textbf{82.1} \\
\midrule
DeDoDe-B + affine augmentation w/o steerer & 40.8 & 57.0 & 70.3 \\
AffEqui-B with pretraining & 47.5 & 63.4 & 75.6 \\
AffSteer-B without pretraining & 47.2 & 63.7 & 76.2 \\
AffEqui-G with pretraining & 51.7 & 68.0 & 80.3 \\
AffSteer-G without pretraining & 50.7 & 67.6 & 80.1 \\
     \bottomrule
     \end{tabular}
     \label{tab:megadepth-ablation}
\end{table}
\section{Ablation study on MegaDepth-1500}
\label{sec:ablation}

We present a couple of ablations on the MegaDepth-1500 test set~\cite{sun2021loftr, li2018megadepth} in Table~\ref{tab:megadepth-ablation}.
First of all, we train DeDoDe-B on MegaDepth with the
same affine augmentations as we train \emph{AffEqui}, but this time without a steerer.
We find that this deteriorates results by a large margin compared to the baseline DeDoDe-B without a steerer.
This corroborates the finding in \cite{bökman2024steerers} that training with large augmentations requires the
addition of a steerer.
Secondly, we train \emph{AffEqui} with homography pretraining and \emph{AffSteer} without homography pretraining and find as explained in the main text that
pretraining matters most for \emph{AffSteer}.

\section{Experiment details}
We use the same hyperparameters as in DeDoDe~\cite{edstedt2024dedode}, with the exception of matching parameters.
Since we use negative L2-distance instead of cosine similarity for measuring description similarity,
the similarity of two descriptions is unbounded from below.
For this reason, we found that lowering the inverse temperature from 20 to 5 was necessary in order to not exaggerate low similarities.

For consistency with \cite{bökman2024steerers}, we use their DeDoDe-C4 detector in our experiments.
An exception from this is the experiment on AIMS,
where we use the DeDoDe-SO2 detector as do \cite{bökman2024steerers}.
For the datasets where there are no reported numbers in \cite{edstedt2024dedode} (\ie Hpatches and WxBS),
we use the DeDoDe-C4 detector for the DeDoDe baseline as well.

When training \emph{AffSteer}, we pretrain for 50k iterations and finetune for 50k iterations, yielding the same total number 100k of training iterations as for \emph{AffEqui} and DeDoDe.

\subsection{Runtime}
Computing similarities multiple (three) times does incur additional runtime---matching for 10k DeDoDe keypoints: 40ms, for AffSteer: 60ms---but this is low compared to the time for detection and description: 240ms for size B and 610ms for G. (On an RTX3080 Ti GPU.)

\section{Using the pipeline to train XFeat} \label{app:xfeat}
XFeat~\cite{potje2024xfeat} is a recent keypoint detector/descriptor which was developed for minimizing computational cost.
To show that the end-to-end pipeline for training upright descriptors that we have proposed does not only work for the DeDoDe architecture, we evaluate using the pipeline for training the descriptor part of XFeat.
\Ie, we pre-train with affine steering and finetune using the max similarity method.
We evaluate with 30k DeDoDe
keypoints on MegaDepth.
For the baseline, we also use 30k DeDoDe
keypoints (giving slightly higher scores than in the original XFeat paper's sparse results, which only used 4k keypoints).
We use LO-RANSAC as in the XFeat paper.
As seen in Table~\ref{tab:xfeat}, a substantial improvement is obtained by training using our pipeline.

\begin{table}[]
    \centering
    \caption{Training the XFeat descriptor in our pipeline improves it significantly.}
    \label{tab:xfeat}
    \begin{tabular}{lrrr}
    \toprule
         & AUC@$5^\circ$ & AUC@$10^\circ$& AUC@$20^\circ$ \\
    \midrule
         XFeat & 43.2 & 57.9 & 69.1 \\
         AffXFeat (ours) & 48.6 & 62.3 & 72.8 \\
     \bottomrule
    \end{tabular}
\end{table}

\end{document}